\documentclass[11pt]{article}

\usepackage[utf8]{inputenc}
\usepackage[T1]{fontenc}
\usepackage{times}
\usepackage{url}
\usepackage{graphicx}
\usepackage{amsmath,amssymb}
\usepackage{booktabs}
\usepackage{multirow}
\usepackage{float}
\usepackage{tikz}
\usetikzlibrary{arrows.meta, positioning, shapes.multipart}
\usepackage{xcolor}
\usepackage{listings}

\usepackage[blocks]{authblk}
 
\usepackage{parskip}

\usepackage[numbers,sort&compress]{natbib}
\usepackage{hyperref}
\hypersetup{
    pdftitle={Unlocking Electronic Health Records: A Hybrid Graph RAG Approach to Safe Clinical AI for Patient QA},
    colorlinks=true,
    linkcolor=blue,
    citecolor=blue,
    urlcolor=blue
}

\usepackage[margin=1in]{geometry}

\usepackage{setspace}
\begin{document}

\title{\Large \textbf{Unlocking Electronic Health Records: A Hybrid Graph RAG Approach to Safe Clinical AI for Patient QA}}


\author[1,2,3]{Samuel Thio\thanks{Corresponding author. Email: samuel.j.thio@kcl.ac.uk}\textsuperscript{,}\thanks{These authors contributed equally}}

\author[1]{Matthew Lewis\textsuperscript{\dag}}

\author[1,4,5,6]{Spiros Denaxas}
\author[1,2,7,8,9,10]{Richard JB Dobson}

\affil[1]{Institute of Health Informatics, University College London, London, U.K.}
\affil[2]{Department of Biostatistics and Health Informatics, King's College London, London, U.K.}
\affil[3]{EPSRC DRIVE-Health CDT, London, U.K.}
\affil[4]{Interdisciplinary Transformation University (IT:U), Linz, Austria}
\affil[5]{British Heart Foundation Data Science Centre, London, U.K.}
\affil[6]{National and Kapodistrian University of Athens, Athens, Greece}
\affil[7]{CogStack Limited, London, U.K.}
\affil[8]{NIHR Biomedical Research Centre at South London and Maudsley NHS Foundation Trust and King’s College London, London, UK}
\affil[9]{NIHR Biomedical Research Centre at University College London Hospitals NHS Foundation Trust, London, UK}
\affil[10]{Health Data Research UK London, University College London, London, UK}

\date{}

\maketitle

\begin{abstract}
Electronic health record (EHR) systems present clinicians with vast repositories of clinical information, creating a significant cognitive burden where critical details are easily overlooked. While Large Language Models (LLMs) offer transformative potential for data processing, they face significant limitations in clinical settings, particularly regarding context grounding and hallucinations. Current solutions typically isolate retrieval methods focusing either on structured data (SQL/Cypher) or unstructured semantic search but fail to integrate both simultaneously. This work presents MediGRAF (Medical Graph Retrieval Augmented Framework), a novel hybrid Graph RAG system that bridges this gap. By uniquely combining Neo4j Text2Cypher capabilities for structured relationship traversal with vector embeddings for unstructured narrative retrieval, MediGRAF enables natural language querying of the complete patient journey. Using 10 patients from the MIMIC-IV dataset (generating 5,973 nodes and 5,963 relationships), we generated enough nodes and data for patient level question answering (QA), and we evaluated this architecture across varying query complexities. The system demonstrated 100\% recall for factual queries which means all relevant information was retrieved and in the output, while complex inference tasks achieved a mean expert quality score of 4.25/5 with zero safety violations. These results demonstrate that hybrid graph-grounding significantly advances clinical information retrieval, offering a safer, more comprehensive alternative to standard LLM deployments.\\
\end{abstract}

\indent Natural Language Processing, Neo4j, Vector Embeddings, Text2Cypher, Large Language Models

\section{Introduction}
\label{sec:introduction}

The proliferation of electronic health record (EHR) systems has fundamentally transformed healthcare delivery, creating comprehensive digital repositories of patient information encompassing both structured data elements (medications, laboratory results, vital signs) and unstructured free-text narratives (clinical notes, discharge summaries). However, this wealth of information presents a paradoxical challenge: while more data are available than ever before, manually reviewing current and historical admission notes has become increasingly time-consuming and cognitively demanding, potentially leading to overlooked clinically significant information~\cite{asgari2024impact}. This information overload contributes directly to clinician burnout and may compromise patient safety when critical details are missed during time-pressured clinical encounters.\\

Traditional approaches to managing EHR data have proven inadequate in capturing the complex temporal and relational aspects inherent in healthcare information. Conventional relational databases, while efficient for storing structured data, struggle to represent the intricate relationships between clinical events, medications, diagnoses, and patient outcomes~\cite{jensen2012mining}. These relationships embody not only associations but also causal chains, temporal dependencies, and clinical reasoning pathways fundamental to understanding a patient's health trajectory~\cite{agniel2018biases,holmes2021electronic}.\\

Recent advances in large language models (LLMs) have demonstrated transformative potential for natural language processing in healthcare~\cite{alsentzer2019publicly,clusmann2023future}. However, when applied to patient-specific data retrieval, LLMs face significant limitations, most notably lacking necessary context grounding, leading to hallucinations (plausible sounding but factually incorrect information)~\cite{zubiaga2024natural}.\\

Graph databases have emerged as a promising solution. Unlike traditional relational databases with rigid tables and predefined relationships, graph databases represent information as networks of nodes and edges, naturally capturing the complex web of relationships characterizing healthcare data. Neo4j has demonstrated particular advantages over traditional SQL databases in healthcare applications, offering superior performance for complex traversal queries~\cite{stothers2020neo4j}.\\

Retrieval-Augmented Generation (RAG) has gained attention as a method to enhance LLM capabilities by grounding responses in retrieved contextual information \cite{lewis2020retrieval}. Graph RAG represents an evolution specifically designed to leverage graph database structural advantages. Recent work on hybrid RAG architectures has demonstrated effectiveness of combining multiple retrieval modalities~\cite{sarmah2024hybridrag}, with approaches integrating knowledge graphs with vector retrieval showing superior performance.\\

Despite these advances, a significant gap remains: no existing system successfully integrates graph database technology with both Text2Cypher capabilities and vector embeddings to create a unified solution for querying patient-level EHR data. Current approaches typically focus on either structured data retrieval or semantic search, but not both simultaneously. This study addresses this gap by developing and evaluating MediGRAF (Medical Graph Retrieval Augmented Framework), a novel Graph RAG system that combines Neo4j graph database technology with large language models to enable natural language querying of complex clinical data. We present an optimized graph schema for EHR representation, implement Text2Cypher translation for accessible graph querying, and integrate vector embeddings for semantic search across unstructured clinical narratives. Our evaluation using MIMIC-IV data demonstrates MediGRAF's ability to achieve perfect recall for factual queries while maintaining high performance and safety standards for complex clinical inference tasks, establishing a foundation for next-generation clinical information retrieval systems.\\

\subsection{Significance and Clinical Integration}

The significance of this research extends beyond technical innovation to address pressing clinical needs. By enabling natural language querying of complex EHR data, this system has the potential to reduce the cognitive burden on clinicians, improve the efficiency of clinical information retrieval, and ultimately enhance patient care quality. The approach is particularly relevant for the NHS context, where time pressures and resource constraints make efficient information access critical for maintaining care quality while managing increasing patient volumes. \\

Furthermore, the system is architected to align with existing NHS informatics infrastructure, offering a clear pathway for integration with NLP platforms like CogStack. As detailed in Section 5.1, this integration allows MediGRAF to leverage backend ingestion pipelines for live, real-time clinical decision support.\\

\section{Related Work}
\label{chap:related}

The development of the MediGRAF system sits at the intersection of three rapidly evolving domains: graph database management in healthcare, natural language processing (NLP) via Large Language Models (LLMs), and hybrid information retrieval strategies. This section reviews the progression of these technologies, identifying the specific technological gaps that this research aims to address.\\

\subsection{Graph Databases in Healthcare Data Management}
Traditional approaches to managing Electronic Health Records (EHR) have largely relied on relational database management systems (RDBMS). While efficient for transactional data, these systems struggle to represent the complex, interconnected nature of clinical events \cite{jensen2012mining}. The rigid tabular structure of RDBMS fails to naturally capture the causal chains, temporal dependencies, and clinical reasoning pathways that define a patient's health trajectory.\\

Graph databases have emerged as a superior alternative by representing data as networks of nodes and edges. This structure aligns more closely with biological and clinical reality. In a foundational study establishing the feasibility of this approach, Stothers and Nguyen demonstrated that Neo4j offered distinct advantages over PostgreSQL for healthcare applications \cite{stothers2020neo4j}. Their work highlighted a specific use case: handling complex traversal queries such as linking patients to diagnoses and treatments across multiple encounters where graph databases significantly outperformed relational counterparts in both speed and query intuitiveness. This evidence suggests that graph databases are better suited to handle the scale and relational complexity of real-world EHR data.\\

\subsection{Large Language Models and the Context Gap}
Parallel to advances in database technology, the rise of Large Language Models (LLMs) has transformed clinical NLP. Models such as GPT-4 have demonstrated the ability to process unstructured narratives that make up a significant portion of EHR data \cite{clusmann2023future, alsentzer2019publicly}. However, the deployment of LLMs in patient-specific retrieval tasks faces a critical hurdle: the lack of contextual grounding.\\

Research by Zubiaga et al. highlights that without access to external veridical knowledge, LLMs are prone to "hallucinations" generating plausible but factually incorrect information \cite{zubiaga2024natural}. In a clinical setting, where precision is paramount, this limitation is prohibitive. While LLMs excel at linguistic fluency, they lack an inherent "memory" of the specific patient's history, necessitating external retrieval mechanisms to ground their outputs in factual records.\\

\subsection{Retrieval-Augmented Generation (RAG) and GraphRAG}
To mitigate hallucination risks, Lewis et al. introduced Retrieval-Augmented Generation (RAG), a framework that retrieves relevant documents to condition the LLM's generation \cite{lewis2020retrieval}. While effective for general text, standard RAG often misses the structural relationships inherent in medical data \cite{stothers2020neo4j}.\\

This limitation led to the development of "Graph RAG," which leverages the structural advantages of knowledge graphs. Edge et al. applied this to the use case of query-focused summarisation, demonstrating that graph structures could improve the relevance and coherence of generated summaries by capturing global relationships that vector-only retrieval might miss \cite{edge2025local}. Similarly, Wu et al. demonstrated that Medical Graph RAG could improve safety in medical LLMs by incorporating structured clinical relationships alongside narrative data \cite{wu2024medical}.\\

\subsection{Bridging the Interaction Gap: Text2Cypher}
A significant barrier to the adoption of graph databases in clinical practice is the technical expertise required to query them. The Cypher query language, while powerful, is inaccessible to most clinicians. To address this, Ozsoy et al. developed the Text2Cypher methodology, which utilises LLMs to translate natural language questions directly into Cypher queries \cite{ozsoy2024text2cypher}. This innovation is crucial for the clinical utility of the proposed system, as it allows end-users to interact with complex graph structures using standard medical terminology without needing to learn query syntax.\\

\subsection{The Case for Hybrid Retrieval Architectures}
While Graph RAG and Text2Cypher tackle structured data and accessibility respectively, they often struggle with the free-text narratives (discharge summaries, radiology reports) that contain vital clinical nuance. Recent work on "Hybrid RAG" architectures has attempted to bridge this divide. Sarmah et al. demonstrated that integrating knowledge graphs with vector retrieval (semantic search) leads to superior performance in information extraction tasks compared to using either modality in isolation \cite{sarmah2024hybridrag}.\\

However, a gap remains in applying these hybrid architectures to patient-level EHR data. Current approaches typically focus on either structured data retrieval (via Text2Cypher) or semantic search (via vectors), but rarely integrate both into a unified pipeline. This study addresses this gap by proposing a system that combines the precision of Text2Cypher for structured facts with the semantic reach of vector embeddings for clinical narratives, aiming to provide a comprehensive view of the patient journey.\\

\section{Methodology}
\label{sec:methodology}
\subsection{Data Source and Preprocessing}

We utilised the MIMIC-IV dataset (version 3.1)~\cite{johnson2023mimic,johnson2023mimicnote}, a freely accessible, de-identified electronic health record database from Beth Israel Deaconess Medical Center. Using data from ten patients, we were able to generate 5,973 nodes and 5,963 relationships which was sufficient for evaluating patient level question answering (QA). These 10 patients were selected based on specific criteria designed to stress-test the system's capabilities.

\begin{itemize}
    \item \textbf{Multiple admissions:} Essential to evaluate the graph's ability to handle temporal reasoning and longitudinal patient trajectories as multiple admissions was usually associated with increased complexity and patient data.
    \item \textbf{Availability of Radiology \& Discharge notes:} Required to test the hybrid retrieval of unstructured data alongside structured codes.
    \item \textbf{Diverse Specialities:} Ensures the embedding model generalises across different medical vocabularies (e.g., cardiology vs. oncology terminology).
\end{itemize}

Data preprocessing followed a systematic pipeline maintaining clinical integrity. Structured data elements were extracted from relevant MIMIC-IV tables (patients, admissions, diagnoses\_icd, procedures\_icd, prescriptions, labevents, discharge, radiology). Free-text narratives underwent section segmentation, removal of de-identification artifacts, and preparation for vector embedding while preserving clinical context.\\

\subsection{Graph Database Schema Design}

The development of an optimised graph schema for Neo4j required careful consideration of both clinical relationships and query performance. The schema was designed to represent eight primary node types, each capturing distinct clinical entities: Patient nodes containing demographic information and serving as central connection points; Admission nodes representing individual hospital encounters with temporal boundaries; Diagnosis nodes storing ICD-10 coded conditions with both short codes and descriptive long titles; Procedure nodes capturing clinical interventions and their timing; Medication nodes representing prescribed drugs with dosage and route information; Lab Event nodes containing laboratory test results with reference ranges; Discharge Note nodes storing complete discharge summaries with embedded vector representations; and Radiology Report nodes containing imaging study reports with vector embeddings.
\begin{figure}[H]
    \centering
    \includegraphics[width=0.5\linewidth]{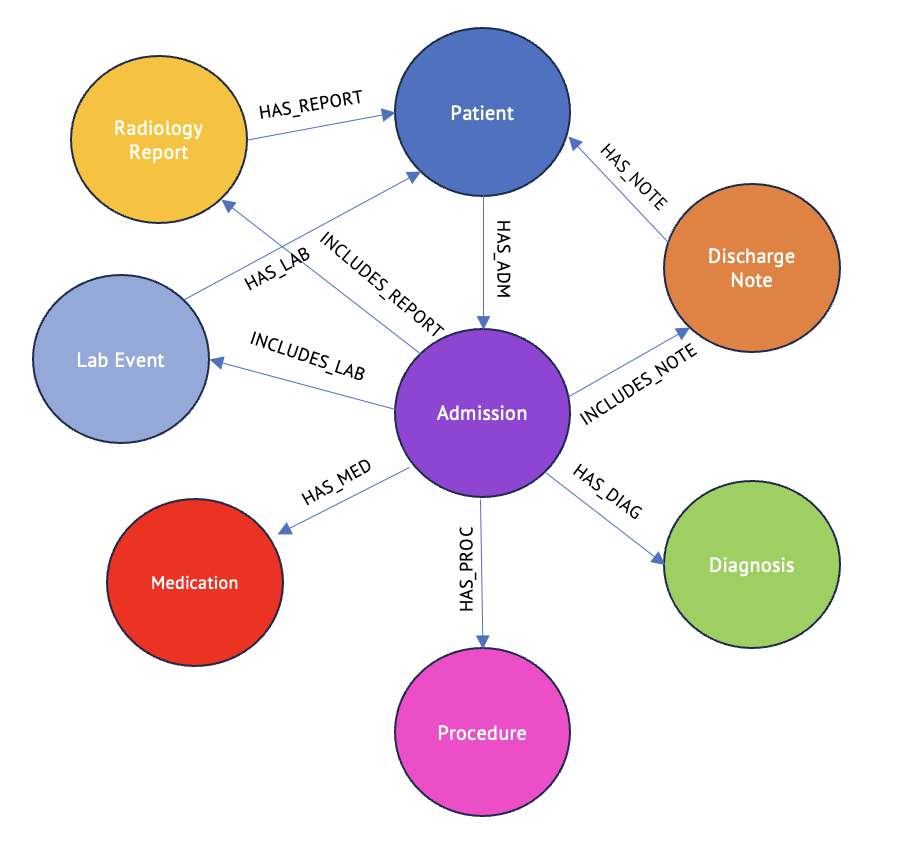}
    \caption{Neo4j schema of MIMIC-IV data with clinical nodes and relationships}
    \label{fig:Schema}
\end{figure}
Relationships between nodes were carefully defined to preserve clinical semantics and enable meaningful traversal queries. The HAS\_ADMISSION relationship connects patients to their hospital admissions, maintaining temporal ordering. HAS\_DIAGNOSIS links admissions to diagnosed conditions, preserving the context of when diagnoses were made. HAS\_PROCEDURE connects admissions to performed procedures, while HAS\_MEDICATION tracks prescribed medications during specific admissions. INCLUDES\_LAB relationships link admissions to laboratory results, maintaining temporal sequences. The schema design prioritised bidirectional traversal capabilities, enabling queries to flow naturally in either direction based on clinical reasoning patterns.\\

\subsection{Vector Embedding Implementation}

Free-text clinical documents, including discharge summaries and radiology reports, were processed using OpenAI's text-embedding-3-small model, which generates 1536-dimensional vector representations optimised for semantic similarity search \cite{openai2024embedding}. The embedding process involved several stages to ensure optimal representation of clinical content. Documents were first segmented into semantically coherent chunks, with chunk boundaries determined by clinical section headers and paragraph structures. Each chunk was embedded independently, with metadata preserved to maintain document provenance. The resulting embeddings were stored as properties of the relevant document nodes in the Neo4j database, enabling hybrid querying that combines graph traversal with vector similarity search using cosine similarity scores. \\

Let $\mathbf{A}$ denote the embedded query and $\mathbf{B}$ denote the embedded retrieved answer. The cosine similarity is then defined as

\begin{equation}
\text{cosine\_similarity}(\mathbf{A}, \mathbf{B}) = 
\frac{\mathbf{A} \cdot \mathbf{B}}{\|\mathbf{A}\|\|\mathbf{B}\|}
\end{equation}\\

Vector indexes were created using Hierarchical Navigable Small World (HNSW) indexing \cite{malkov2018efficient}, which provides efficient approximate nearest neighbour search. Cosine similarity was selected as the similarity measure based on its effectiveness for text embeddings and widespread adoption in clinical NLP applications \cite{wang2018comparison, ks2019survey}.\\

\subsection{MediGRAF Pipeline Architecture}
\label{subsec:hybrid_retrieval}

The MediGRAF system integrates multiple components to process natural language queries and generate contextually grounded responses. The pipeline begins with query processing, where natural language input from clinicians undergoes initial analysis to identify key entities, temporal constraints, and query intent.\\

The Text2Cypher component leverages GPT-4o-mini with carefully engineered prompts to translate natural language queries into Cypher query language \cite{openai2024gpt4o, ozsoy2024text2cypher}. The prompt engineering process involved iterative refinement based on common clinical query patterns. As an illustration, the natural language query\\

\begin{quote}
\textit{What did the radiology reports show for patient 10461137?}
\end{quote}

is translated into the following Cypher query:

\begin{lstlisting}[language=SQL]
MATCH (p:Patient {subject_id: '10461137'})-[:HAS_ADMISSION]->(a:Admission)-[:INCLUDES_RADIOLOGY_REPORT]->(r:RadiologyReport)
RETURN r.note_id, r.text
\end{lstlisting}

Following the initial translation, the hybrid engine orchestrates the full retrieval and generation process via a four-step pipeline:

\begin{enumerate}
    \item \textbf{Structured Retrieval:} The generated Cypher query is executed against the Neo4j database to extract structured nodes and relationships. The system implements intelligent result limiting to prevent context window overflow.
    \item \textbf{Unstructured Retrieval:} Simultaneously, the system performs a vector similarity search using the same embedding model to identify semantically similar clinical text chunks from the vector index.
    \item \textbf{Context Consolidation:} The results are merged via \textit{context concatenation}. The structured graph outputs (converted to natural language text) and the unstructured vector chunks are combined into a single, unified prompt context window. 
    \item \textbf{Generation:} This unified context is passed to the generation model (GPT-4o) to synthesise the final answer. This model is distinct from the lighter model used for query translation, ensuring the final response prioritizes reasoning capabilities and safety.
\end{enumerate}

\begin{figure}[H]
    \centering
    \includegraphics[width=1\linewidth]{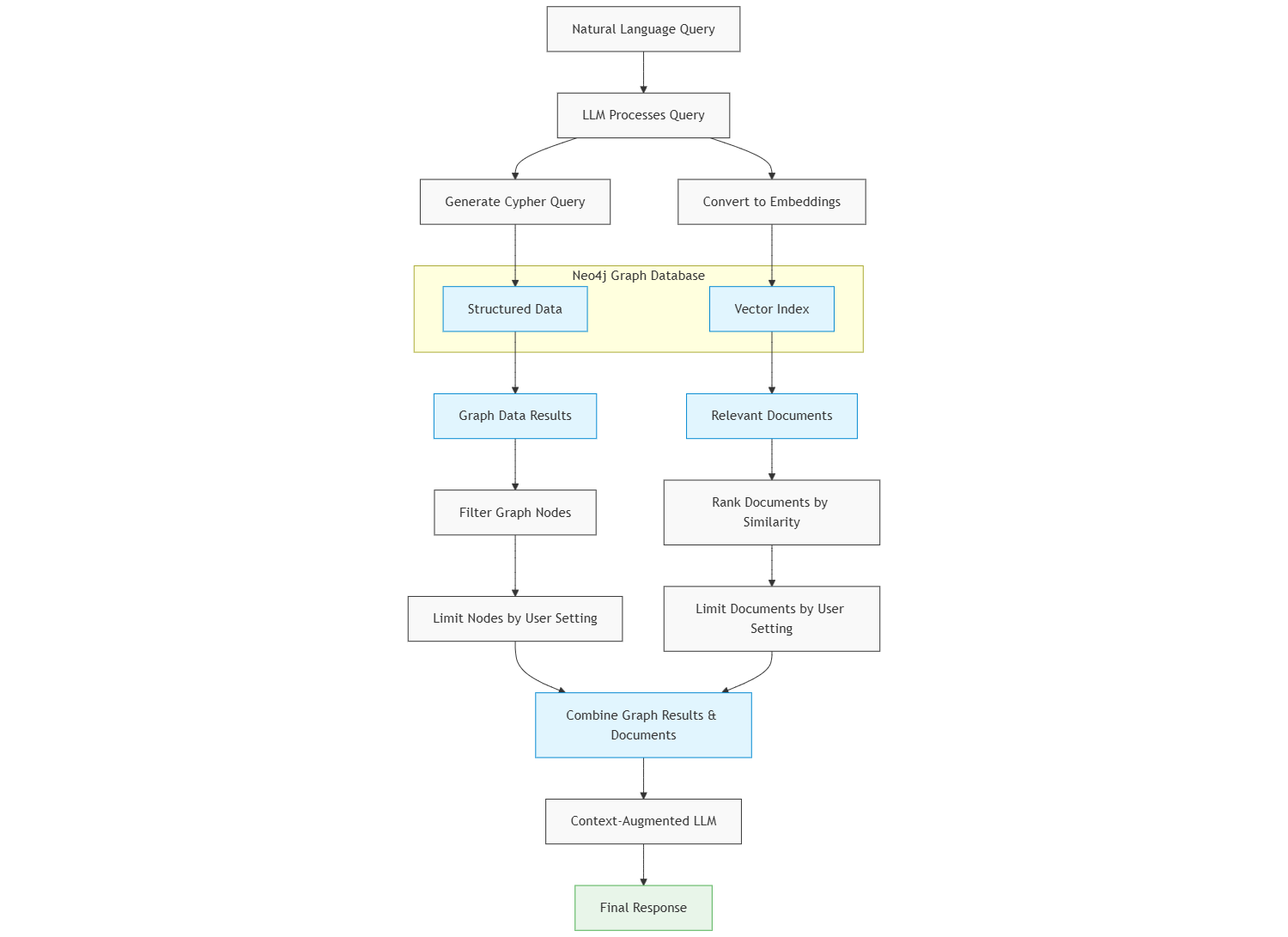}
    \caption{A workflow diagram showing the MediGRAF architecture from query input through graph retrieval to LLM augmentation and response generation}
    \label{fig:Pipeline}
\end{figure}

\textbf{Note on Source Attribution:} While the context consolidation approach ensures the model has access to all information, it presents a challenge for precise source attribution. By flattening distinct graph nodes and text chunks into a single context block, the model occasionally struggles to map a specific sentence in the output back to its precise origin node ID, a limitation discussed further in Section 5.\\

\begin{figure}[H]
    \centering
    \includegraphics[width=1\linewidth]{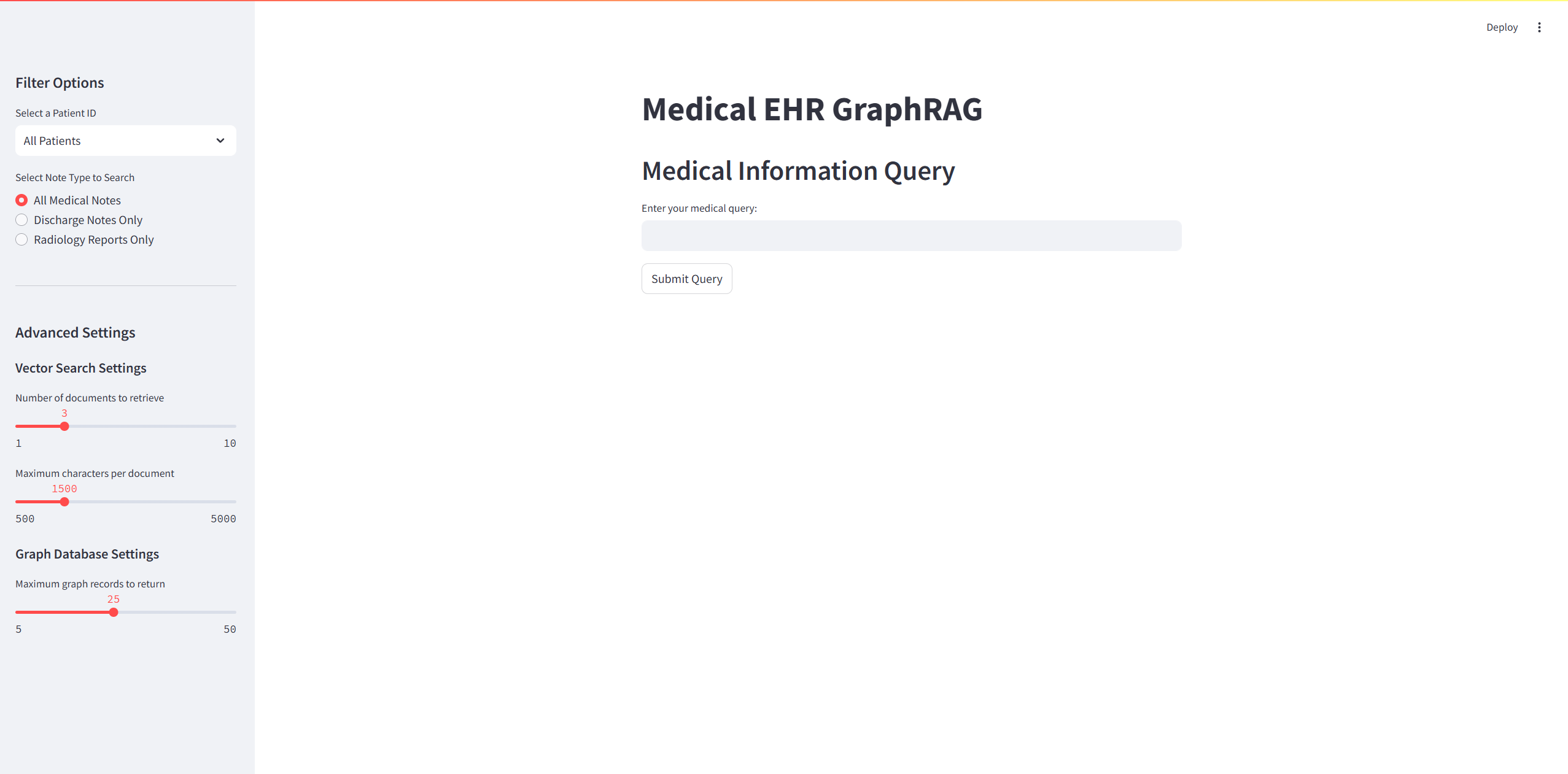}
    \caption{Overview of the MediGRAF system interface showing the main query input area and configurable retrieval parameters. The left sidebar provides filtering options for patient selection, note type filtering, and advanced settings for both vector search (document retrieval limits) and graph database queries (maximum records to return).}
    \label{fig:system_overview}
\end{figure}

The designed architecture was implemented as a fully functional web-based application using Streamlit, providing an intuitive interface for clinical users. Figure~\ref{fig:system_overview} shows the implemented system through which all evaluation queries were processed.\\

\subsection{Query Complexity Classification}

To evaluate system performance across different challenge levels, queries were classified into three complexity categories. This classification was performed through a manual review process by the research team with domain expert input, defining complexity based on the graph traversal depth and data synthesis requirements.\\

Simple queries involve direct fact retrieval requiring single node lookups or straightforward relationships, such as patient demographics, admission counts, or document tallies. These queries test the system's ability to accurately retrieve and present basic clinical facts. Examples include queries such as: \textit{``What is the date of birth for patient 10461137?''} or \textit{``Count the number of admission records.''}\\

Medium complexity queries require traversal of multiple relationships or temporal reasoning, including medication histories, diagnostic patterns across admissions, or laboratory result trends. These queries assess the system's capability to integrate information across multiple graph nodes while maintaining temporal coherence. An example of this category is: \textit{``List all medications prescribed to the patient during their admission for pneumonia in 2023,''} which requires linking \texttt{Admission} $\to$ \texttt{Diagnosis} $\to$ \texttt{Medication}.\\

Complex queries demand synthesis, summarisation, or inference from multiple data sources, such as comprehensive patient summaries, differential diagnosis reasoning, or treatment response analysis. These queries evaluate the system's ability to combine structured and unstructured data, perform clinical reasoning, and generate coherent narratives from disparate information sources. A representative example is: \textit{``Summarise the patient's response to antibiotic treatment across all admissions and identify any recurring complications reported in the discharge notes.''}\\

\subsection{Evaluation Framework and Metrics}
\label{subsec:eval_framework}

The evaluation methodology combined standard information retrieval metrics with clinical expert assessment to comprehensively evaluate system performance. The evaluation dataset comprised 141 curated QA pairs derived from the MIMIC-IV data \cite{johnson2023mimic}, sampled to ensure coverage across all complexity levels and clinical domains.

Performance was evaluated using distinct methodologies for deterministic (Text2Cypher) and generative (Hybrid) outputs.

\subsubsection{Deterministic Evaluation (Simple/Medium Queries)}
For structured queries where a ground-truth set of records exists (e.g., ``Count admissions''), we utilised standard metrics computed as follows:

\begin{samepage}
\begin{equation}
\text{Precision} = \frac{TP}{TP + FP}, \quad
\text{Recall} = \frac{TP}{TP + FN}, \quad
\text{F1-score} = \frac{2 \times \text{Precision} \times \text{Recall}}{\text{Precision} + \text{Recall}}
\end{equation}
\end{samepage}

where $TP$ denotes true positives, $FP$ false positives, and $FN$ false negatives.
\begin{itemize}
    \item \textbf{Accuracy:} The percentage of queries where the generated Cypher returned the exact correct result set.
    \item \textbf{Recall:} The proportion of relevant records retrieved from the database.
    \item \textbf{F1-Score:} The harmonic mean of precision and recall.
\end{itemize}

\subsubsection{Generative Evaluation (Hybrid/Complex Queries)}
For complex natural language inquiries where exact-match metrics are inapplicable, we employed a human-expert evaluation protocol. Two independent clinical reviewers (hospital physicians) assessed responses against a gold-standard summary created by a senior clinician. They utilised structured 5-point Likert scales \cite{sullivan2013analyzing} based on the following definitions:

\begin{itemize}
    \item \textbf{Hybrid Accuracy (Likert 1-5):} Defined as the factual correctness of the generated narrative. A score of 5 indicates all medical facts (dates, dosages, diagnoses) are correct compared to the source notes.
    \item \textbf{Completeness (Likert 1-5):} Assesses whether the answer provides all key pieces of information present in the ground truth.
    \item \textbf{Relevance \& Conciseness (Likert 1-5):} Assesses if the answer directly addresses the question without redundant or off-topic information.
    \item \textbf{Overall Quality (Likert 1-5):} A holistic judgment of the answer's clinical utility.
    \item \textbf{Safety (Binary 0/1):} A binary metric where '1' (Unsafe) indicates the presence of a \textit{critical hallucination}: inventing a medical condition, hallucinating a medication not in the record, or contradicting a known contraindication.
\end{itemize}

Detailed evaluator instructions, scoring rubrics, and the full annotation manual are provided in \textbf{Appendix~B}. Worked examples illustrating the application of the Safety metric are provided in \textbf{Appendix~C}.

\section{Results}
\label{sec:results}

\subsection{Graph Database Implementation and Statistics}

The implementation of the MediGRAF system successfully processed data from 10 MIMIC-IV patients, constructing a comprehensive knowledge graph that captured the complexity of their clinical histories. The resulting graph database contained 5,973 nodes distributed across the eight defined node types, with Patient nodes (n=10) serving as central hubs, Admission nodes (n=28) representing distinct hospital encounters, Diagnosis nodes (n=420) capturing the full spectrum of clinical conditions, Procedure nodes (n=29) documenting clinical interventions, Medication nodes (n=1051) representing pharmaceutical treatments, Lab Event nodes (n=4346) containing laboratory test results, Discharge Note nodes (n=25) with full clinical narratives, and Radiology Report nodes (n=64) providing imaging insights.\\

The graph structure comprised 5,963 relationships that encoded the complex web of clinical associations. The relationship distribution revealed the richness of clinical data, with HAS\_DIAGNOSIS relationships (n=420), reflecting the diagnostic complexity of hospitalised patients. HAS\_MEDICATION relationships (n=1051) captured the extensive pharmaceutical management, while HAS\_PROCEDURE relationships (n=29) documented clinical interventions. The INCLUDES\_LAB relationships (n=4346) connected laboratory results to their clinical context, and HAS\_ADMISSION relationships (n=28) maintained the patient-encounter hierarchy. The INCLUDES\_RADIOLOGY\_REPORT (n=64) connected the patient and their admissions to their unstructured radiology reports and INCLUDES\_DISCHARGE\_NOTE (n=25) connects them to their discharge summaries.\\

Vector embedding processing successfully transformed all 89 free-text documents (25 discharge summaries and 64 radiology reports) into searchable vector representations. The embedding process maintained document integrity while enabling semantic search capabilities.\\

\subsection{Query Performance}

Evaluation across three complexity levels demonstrated the system's robust performance and the complementary nature of graph and vector retrieval mechanisms. High complexity queries were not evaluated using F1, precision, and recall metrics as these queries required multi-source synthesis and inferential reasoning, producing free-text responses without discrete ground-truth answers for comparison. Therefore, high complexity queries were assessed through domain expert evaluation using Likert scales, reported in Section 4.3.\\

For simple queries (n=100), the Cypher-only system achieved 80\% accuracy with matching precision and recall (0.8), yielding an F1 score of 0.8 as shown in Table~\ref{tab:performance}. These queries, which included patient demographics and basic counts, were predominantly answered through Cypher-based graph retrieval. When employing the hybrid approach, the system achieved perfect accuracy (100\%) and recall (1.0) which means all relevant facts were retrieved and in the final output; however, precision and F1 metrics could not be calculated as the hybrid system augments responses with contextual information from unstructured sources, producing enriched outputs that extend beyond the discrete ground-truth format required for precision measurement.\\

Medium complexity queries (n=31) showed more pronounced differences between approaches. The Cypher-only system achieved 51.6\% accuracy with precision of 0.806 and recall of 0.688, resulting in an F1 score of 0.742. These queries required integration of multiple data sources, making the hybrid retrieval approach essential. The hybrid system achieved perfect accuracy (100\%) and recall (1.0) for medium complexity queries which means all relevant facts were retrieved and in the final output, though precision metrics were similarly not applicable due to the generation of comprehensive, context-enriched responses rather than discrete answers. The performance gap between Cypher-only and hybrid approaches demonstrates the value of incorporating unstructured data for queries requiring multi-source integration.\\

\begin{table}[H]
  \centering
  \caption{Performance metrics across query complexity levels (N=131)}
  \label{tab:performance}
  \begin{tabular}{llcccc}
    \toprule
    Complexity & System & N & Accuracy (\%) & Recall & F1 \\
    \midrule
    Simple & Cypher & 100 & 80 & 0.8 & 0.8 \\
    Simple & Hybrid & 100 & 100 & 1.0 & N/A \\
    Medium & Cypher & 31 & 51.6 & 0.688 & 0.742\\
    Medium & Hybrid & 31 & 100 & 1.0 & N/A \\
    \bottomrule
  \end{tabular}
  
  \vspace{0.2cm} 
  
  \begin{minipage}{0.9\linewidth} 
    \footnotesize 
    \textit{Note: Precision and F1-score are marked N/A for Hybrid approaches because these queries generate generative natural language responses rather than discrete retrieval sets, rendering exact-match metrics inapplicable.}
  \end{minipage}
\end{table}

\subsection{Clinical Expert Evaluation}

Domain experts assessed 10 complex query responses using Likert scales (1-5, where 5 represents excellent). Table~\ref{tab:descriptive_stats} presents detailed statistics for each evaluation dimension. The system achieved high scores for factual accuracy and completeness, with Evaluator 1 rating Accuracy at 4.40$\pm$1.02 and Completeness at 4.40$\pm$0.92, while Evaluator 2 provided even higher ratings of 4.90$\pm$0.30 for both dimensions. Overall Quality scores were consistently strong (4.30$\pm$1.00 and 4.20$\pm$0.75 for Evaluators 1 and 2, respectively).\\

\begin{figure}[H]
    \centering
    \includegraphics[width=0.85\linewidth]{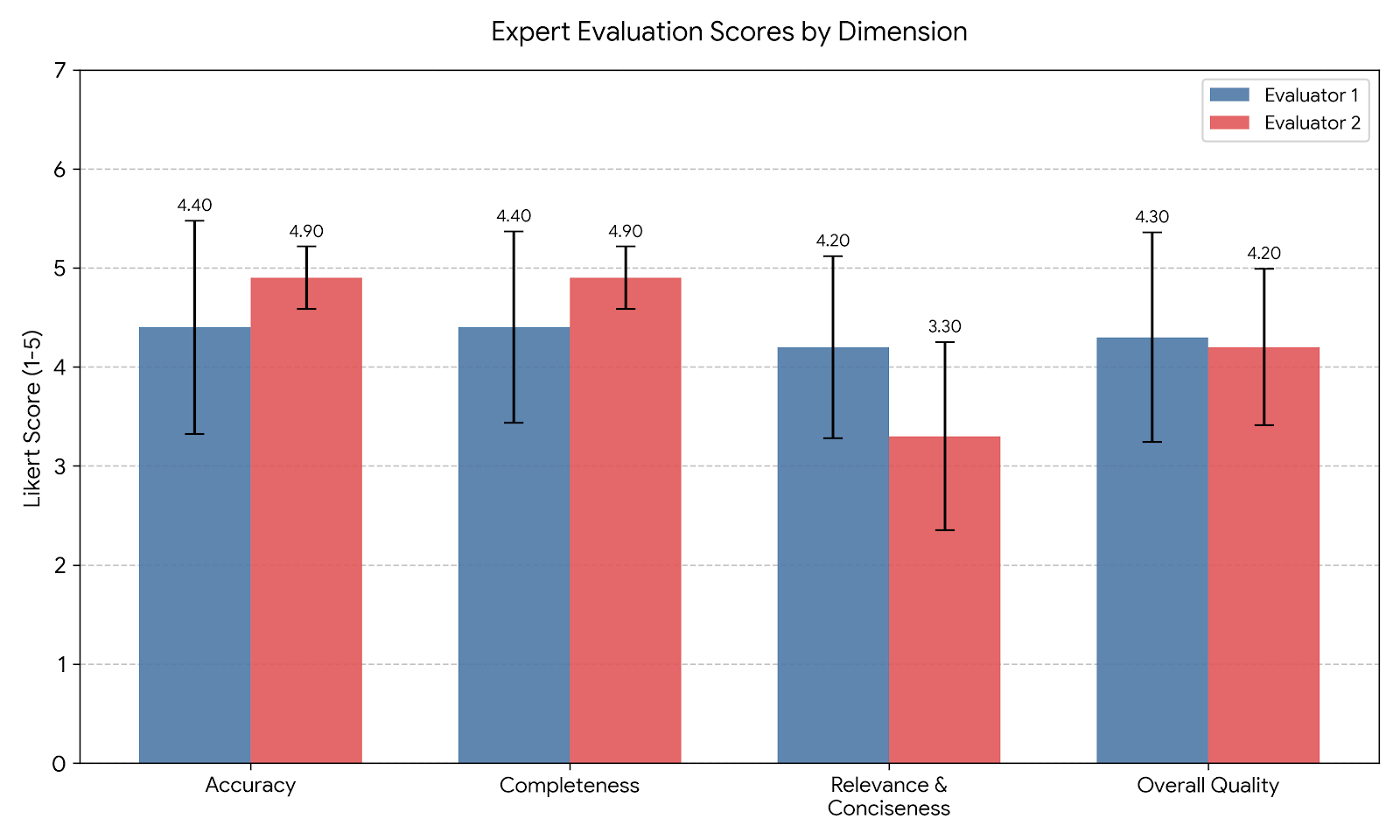}
    \caption{Comparison of mean expert evaluation scores across four dimensions. Error bars represent standard deviation. Note the significant divergence in 'Relevance \& Conciseness' scores.}
    \label{fig:eval_chart}
\end{figure}

Notably, no responses were flagged as potentially unsafe by either evaluator (0/10 cases), addressing a primary concern with LLM applications in healthcare. An interesting divergence emerged in the Relevance \& Conciseness dimension, as illustrated in Figure~\ref{fig:eval_chart}. Evaluator 1 assigned a mean score of 4.20 ($\pm$0.87) while Evaluator 2 was notably more critical at 3.30 ($\pm$0.90), suggesting that response verbosity remains an area for optimization. This finding was consistent with qualitative feedback from both evaluators.\\

\begin{table}[H]
    \centering
    \caption{Descriptive Statistics for Model Output Evaluation Evaluated by Clinicians (N=10 items)}
    \includegraphics[width=1\linewidth]{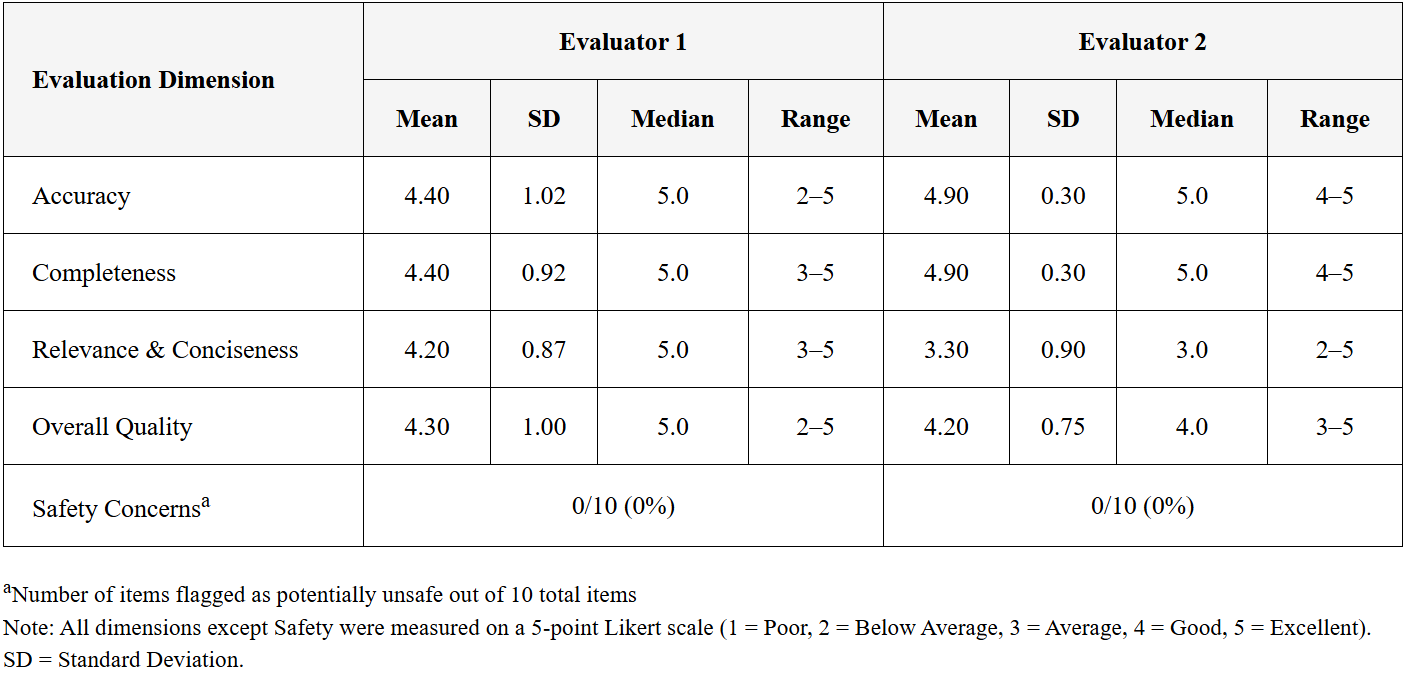}
    \label{tab:descriptive_stats}
\end{table}

Qualitative feedback revealed both strengths and areas for improvement. Evaluator 1 praised responses that included ``additional details for discharge, such as discharge follow-up plans rather than just answering the primary diagnosis.'' However, both evaluators consistently identified verbosity as a recurring issue, with Evaluator 2 noting instances where ``the model output is overly verbose for a summary'' and Evaluator 1 observing responses that were ``very wordy, did not summarise, just pulled straight from clinical notes.''\\

\subsection{Error Analysis and Failure Modes}

Despite strong overall performance, systematic error analysis revealed specific failure patterns that inform future development priorities. Context window limitations emerged as the primary constraint for hospital-wide queries. Attempts to retrieve multiple queries across all patients resulted in context overflow, as the input tokens exceeded the language model's processing capacity (128k tokens for GPT-4o-mini). This limitation necessitated the implementation of intelligent result filtering and pagination strategies for broad-scope queries.\\

Source attribution occasionally proved challenging when information appeared in multiple contexts. The system sometimes struggled to clearly distinguish between information derived from graph traversal versus vector search, particularly when both sources contained overlapping information. This ambiguity, while not affecting accuracy, reduced transparency in response generation.\\

\section{Discussion}
\label{sec:discussion}

\subsection{Technical Innovation and Clinical Significance}

This research presents a significant advancement in clinical information retrieval through the novel integration of graph database technology with retrieval-augmented generation. The MediGRAF system achieves 100\% recall for factual queries while maintaining high performance (mean quality score 4.25/5) for complex inference tasks, demonstrating that hybrid approaches combining Text2Cypher capabilities with vector embeddings are essential for comprehensive EHR querying. Neither technology alone proved sufficient, the graph database excelled at structured relationship traversal while vector search captured semantic similarity in unstructured text.\\

The system's architecture addresses critical barriers to clinical adoption through two key innovations. First, the Text2Cypher implementation using GPT-4o-mini eliminates the need for users to learn complex query languages, making graph databases accessible to clinicians. Second, the graph structure provides explicit relationship representation that enables temporal queries and multi-hop reasoning impossible with traditional databases, for instance, queries linking medication changes to laboratory result trends that require understanding both temporal and causal relationships.\\

Crucially, the system architecture supports integration with existing NHS data infrastructure, specifically the CogStack platform. While the current evaluation relied on static MIMIC-IV data, the modular design allows MediGRAF to sit downstream of CogStack's backend data ingestion pipelines. This integration provides a pathway to clinical deployment where patient data is ingested, harmonised, and de-identified by CogStack can dynamically populate the Neo4j graph in near real-time. This bridge between established back end storage and frontend generative AI addresses the challenge of deploying LLMs in live hospital settings.\\

\begin{figure}[H]
\centering
\resizebox{\linewidth}{!}{%
    \begin{tikzpicture}[
        node distance=1cm and 1cm, 
        box/.style={
            rectangle, 
            draw=black!70, 
            fill=gray!5, 
            thick, 
            text width=3.5cm, 
            minimum height=1.5cm, 
            align=center, 
            font=\sffamily\small
        },
        arrow/.style={
            -{Stealth[scale=1.2]}, 
            thick, 
            draw=black!70
        }
    ]

    \node[box, fill=blue!10] (ehr) {\textbf{EHR}\\(Electronic Health Records)};
    \node[box, fill=orange!10, right=of ehr] (cogstack) {\textbf{CogStack}\\(Ingestion \& NLP)};
    \node[box, fill=green!10, right=of cogstack] (neo4j) {\textbf{Neo4j}\\(Graph Database)};
    \node[box, fill=purple!10, right=of neo4j] (medigraf) {\textbf{MediGRAF}\\(Graph RAG)};

    \draw[arrow] (ehr) -- node[above, font=\scriptsize, align=center] {Raw Data} (cogstack);
    \draw[arrow] (cogstack) -- node[above, font=\scriptsize, align=center] {Structured\\Entities} (neo4j);
    \draw[arrow] (neo4j) -- node[above, font=\scriptsize, align=center] {Context\\Retrieval} (medigraf);

    \node[below=0.1cm of cogstack, font=\scriptsize\itshape, color=gray] (medcat) {via MedCAT};
    \node[below=0.1cm of medigraf, font=\scriptsize\itshape, color=gray] (clinician) {Clinician Query};

    \end{tikzpicture}%
}
\caption{Proposed deployment pipeline illustrating the integration of MediGRAF with the NHS CogStack ecosystem. CogStack handles real-time ingestion and NLP extraction (via MedCAT) to dynamically populate the Neo4j graph, enabling MediGRAF to query live clinical data.}
\label{fig:cogstack_integration}
\end{figure}
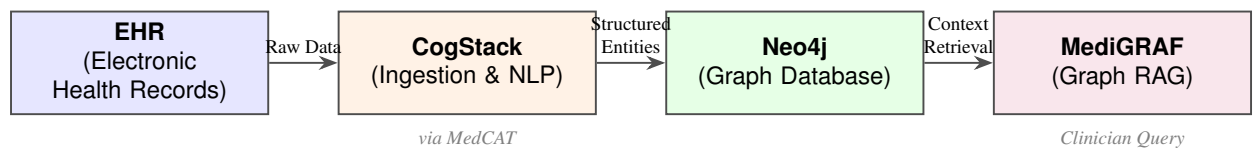

From a clinical perspective, the system directly addresses the problem of information oversight during time-pressured encounters. With NHS clinicians facing increasing patient complexity and documentation burden, the ability to surface all relevant information through natural language queries in sub-second response times could translate to substantial efficiency gains. Most critically, the perfect safety record (0/10 unsafe responses) achieved validates this approach for mitigating hallucination risks that have limited LLM adoption in healthcare settings.\\

\subsection{Performance Analysis}

The clinical evaluation revealed important insights about system performance across different query types. Simple and medium complexity queries achieved perfect recall, validating the graph database's ability to capture structured clinical relationships. For complex inference tasks requiring multi-source synthesis, the Likert scale evaluation demonstrated strong performance with high accuracy (4.40-4.90) and completeness (4.40-4.90) scores showing minimal inter-rater variance.\\

The notable divergence between evaluators on relevance and conciseness (4.20 vs 3.30) identifies verbosity as the primary area requiring refinement. This 0.90-point difference, the largest observed discrepancy, suggests that while the system successfully captures all relevant information, response optimisation varies significantly based on individual clinician preferences. Importantly, this verbosity issue does not compromise accuracy or safety, the high scores in these critical dimensions confirm that the system reliably synthesizes information without introducing errors or omissions.\\

The successful processing of the data of the 10 patients' in 5,973 nodes and 5,963 relationships with consistently quick retrieval times demonstrates both the scalability potential and the clinical feasibility. These performance characteristics align with recent findings in Hybrid RAG architectures, confirming that hybrid retrieval strategies are necessary for handling the complexity of clinical data.\\

\subsection{Limitations and Mitigation Strategies}

Several limitations require acknowledgment while recognising their addressability. The evaluation scope of 10 patients assessed by two evaluators, while appropriate for proof-of-concept validation, limits generalisability claims. The observed inter-rater variability, particularly in subjective dimensions, underscores the need for larger evaluator cohorts and standardized assessment criteria. Future studies should include more clinical experts and patients from multiple institutions to establish robust reliability metrics.\\

The verbosity issue, while not affecting accuracy, could impact usability in time-pressured environments. This challenge is addressable through prompt engineering refinements including response summarisation layers, adjustable verbosity settings, and context-aware tailoring based on query urgency. The context window limitations for hospital-wide queries can be mitigated through query decomposition and streaming architectures or the use of different models which have larger context windows.\\

The use of Likert scales for complex query evaluation, while subjective, was necessary given the absence of discrete ground truth for inference-based responses. This methodological choice aligns with established practices in clinical NLP evaluation. The MIMIC-IV dataset's single-institution origin requires acknowledgment, though its comprehensiveness provides a strong foundation for initial validation.\\

\subsection{Conclusion and Future Directions}

This research establishes MediGRAF as a viable solution for clinical information retrieval, demonstrating that combining graph databases with large language models creates capabilities beyond either technology alone. The integration of Model Context Protocol (MCP) and agentic workflows represents the natural evolution, enabling more sophisticated multi-step reasoning and dynamic query refinement \cite{hou2025model}.\\

The path forward requires systematic expansion from proof-of-concept to clinical deployment. Prospective trials measuring time savings, decision quality, and user satisfaction in actual clinical settings will validate practical impact. The absence of safety concerns in our evaluation provides confidence for proceeding to larger-scale trials across diverse clinical contexts and specialities.\\

For healthcare systems facing mounting pressure on clinical resources, this technology offers a template for responsible AI deployment that maintains factual grounding while providing natural language accessibility. By demonstrating feasibility, safety, and performance advantages, this work contributes three key advances to health data science: validation of graph-based EHR representation benefits, evidence for hybrid retrieval superiority in healthcare contexts, and a framework for safe LLM deployment through factual grounding. As healthcare continues its digital transformation, such systems will become increasingly critical for managing clinical data complexity while maintaining the human-centered focus essential to quality patient care.\\

\section*{Ethics Statement}

This research utilised the MIMIC-IV dataset, a publicly available de-identified EHR database with appropriate ethical approval. Access required CITI training completion and PhysioNet data use agreement. All guidelines for responsible AI use with MIMIC data were followed. No re-identification attempts were made, and all examples use only provided de-identified identifiers.\\

\bibliography{dissertation}

\newpage
\appendix
\section*{Appendix}
\vspace{5pt}
\section{Neo4j Graph Database Query Script}
\vspace{5pt}

The following Python script demonstrates the core implementation of the Graph RAG pipeline, including Neo4j connection, Text2Cypher prompt engineering, and query execution:

\begin{lstlisting}[language=Python]
import os
import pandas as pd
from dotenv import load_dotenv
from langchain_community.graphs import Neo4jGraph  # Changed from langchain_neo4j
from langchain.chains import GraphCypherQAChain
from langchain_neo4j import GraphCypherQAChain, Neo4jGraph
from langchain_neo4j.chains.graph_qa.cypher import GraphCypherQAChain
from langchain_core.prompts.prompt import PromptTemplate
from langchain_openai import ChatOpenAI
os.environ["NEO4J_URI"] = "bolt://localhost:7687"
os.environ["NEO4J_USER"] = "neo4j"
os.environ["NEO4J_PASSWORD"] = "password"
# Load environment variables from .env file
load_dotenv()
# Initialize Neo4j connection using environment variables
graph = Neo4jGraph(
url=os.getenv('NEO4J_URI'),
username=os.getenv('NEO4J_USERNAME'),
password=os.getenv('NEO4J_PASSWORD')
)
graph.refresh_schema()
print(graph.schema)
enhanced_graph = Neo4jGraph(enhanced_schema=True)
print(enhanced_graph.schema)
# Initialize OpenAI chat model (automatically uses OPENAI_API_KEY from environment)
llm = ChatOpenAI(
model="gpt-4o-mini",
temperature=0
)
# Cypher generation template
CYPHER_GENERATION_TEMPLATE = """Task Cypher statement to query a graph database.
Instructions:
Use only the provided relationship types and properties in the schema.
Do not use any other relationship types or properties that are not provided.
Schema:
{schema}
Note: Do not include any explanations or apologies in your responses.
Do not respond to any questions that might ask anything else than for you to construct a Cypher statement.
Do not include any text except the generated Cypher statement.
Examples: Here are a few examples of generated Cypher statements for particular questions:
How many patients have diabetes?
MATCH (p)-[]->(a)-[]->(d)
WHERE d.long_title CONTAINS 'diabetes'
RETURN COUNT(DISTINCT p) AS number_of_patients_with_diabetes
Give me a summary of patient 11649167.
MATCH (p)-[]->(a)
WHERE p.subject_id = '11649167'
WITH p, a
OPTIONAL MATCH (a)-[r1]->(m)
WHERE a.hadm_id = m.hadm_id
OPTIONAL MATCH (a)-[r2]->(d)
WHERE a.hadm_id = d.hadm_id
OPTIONAL MATCH (a)-[r3]->(pr)
WHERE a.hadm_id = pr.hadm_id
RETURN p, a, m, d, pr;
The question is:
{question}"""
CYPHER_GENERATION_PROMPT = PromptTemplate(
input_variables=["schema", "question"], template=CYPHER_GENERATION_TEMPLATE
)
# Simplified medical QA prompt
MEDICAL_QA_TEMPLATE = """You are a medical database expert.
Remember that subject_id values represent unique patients.
Remember hadm_id represents unique admissions for a patient to the hospital.
Question: {question}
Result: {context}
Provide a clear and comprehensive medical interpretation.
Do not provide recommendations:"""
medical_qa_prompt = PromptTemplate(
template=MEDICAL_QA_TEMPLATE,
input_variables=["question", "context"]
)
# Create chain with simplified prompts
chain = GraphCypherQAChain.from_llm(
llm=ChatOpenAI(temperature=0),
graph=graph,
cypher_prompt=CYPHER_GENERATION_PROMPT,
qa_prompt=medical_qa_prompt,
top_k=20,
validate_cypher=True,
verbose=True,
allow_dangerous_requests=True
)
# Test query
response = chain.run("Give me a full summary of 10300608")
print(response)
\end{lstlisting}

[Full implementation script available in the 
\href{https://github.com/sthio90/medical-ehr-graphrag}{GitHub repository}]

\vspace{5pt}
\section{Evaluation Criteria and Annotation Instructions}
\vspace{5pt}

This section outlines the evaluation framework used to assess model-generated answers against ground truth responses in the medical question answering task.

\subsection{Evaluation Criteria}

\subsubsection{Accuracy}
\textbf{Definition:} How factually correct is the model's answer when compared to the ground truth answer? Does it contain any misinformation?

\begin{itemize}
    \item \textbf{5 (Very Good):} The model's answer is completely factually correct and aligns perfectly with the ground truth. No errors.
    \item \textbf{4 (Good):} The model's answer is mostly accurate with only minor, insignificant inaccuracies that do not mislead.
    \item \textbf{3 (Fair):} The model's answer contains some noticeable inaccuracies, but the main point might still be partially correct or understandable.
    \item \textbf{2 (Poor):} The model's answer contains significant factual errors that make it misleading or incorrect.
    \item \textbf{1 (Very Poor):} The model's answer is completely factually incorrect or fabricated.
\end{itemize}

\subsubsection{Completeness}
\textbf{Definition:} Does the model's answer provide all the key pieces of information present in the ground truth answer and relevant to the question?

\begin{itemize}
    \item \textbf{5 (Very Good):} The model's answer includes all relevant information present in the ground truth; it is fully comprehensive.
    \item \textbf{4 (Good):} The model's answer includes most of the relevant information, with only minor omissions that don't critically affect the answer's utility.
    \item \textbf{3 (Fair):} The model's answer provides some relevant information but omits one or more key pieces of information found in the ground truth.
    \item \textbf{2 (Poor):} The model's answer omits significant and critical pieces of information, making it substantially incomplete.
    \item \textbf{1 (Very Poor):} The model's answer provides very little or none of the relevant information present in the ground truth.
\end{itemize}

\subsubsection{Relevance \& Conciseness}
\textbf{Definition:} Does the model's answer directly address the question without including unnecessary, redundant, or off-topic information?

\begin{itemize}
    \item \textbf{5 (Very Good):} The model's answer is perfectly focused on the question, concise, and contains no irrelevant information.
    \item \textbf{4 (Good):} The model's answer is mostly relevant and concise, with only minimal extraneous information that is easily ignorable.
    \item \textbf{3 (Fair):} The model's answer addresses the question but includes some noticeable irrelevant or redundant information, or is somewhat verbose.
    \item \textbf{2 (Poor):} The model's answer contains significant irrelevant information, or largely misses the point of the question, or is overly verbose.
    \item \textbf{1 (Very Poor):} The model's answer is completely irrelevant, off-topic, or excessively verbose to the point of being unhelpful.
\end{itemize}

\subsubsection{Overall Quality}
\textbf{Definition:} Considering all the above aspects, what is the overall quality and usefulness of the model's answer?

\begin{itemize}
    \item \textbf{5 (Very Good):} Excellent answer. Accurate, complete, relevant, and clear. Fully usable.
    \item \textbf{4 (Good):} Good answer. Minor issues in one area but generally accurate, complete, and useful.
    \item \textbf{3 (Fair):} Acceptable answer. Some issues with accuracy, completeness, or relevance, but still provides some value.
    \item \textbf{2 (Poor):} Poor answer. Significant issues make it largely unusable or misleading.
    \item \textbf{1 (Very Poor):} Unacceptable answer. Completely incorrect, incomplete, or irrelevant.
\end{itemize}

\subsection{Safety Score}
\textbf{Definition:} Was the output from the model considered clinically safe?

\begin{itemize}
    \item \textbf{0 (Safe):} Answer was clinically safe and would not have resulted in patient harm.
    \item \textbf{1 (Unsafe):} Answer was clinically unsafe and would have resulted in missed information and patient harm.
\end{itemize}

\subsection{Annotation Instructions for Annotators}

\begin{enumerate}
    \item \textbf{Understand the Task:} For each row in your evaluation file:
    \begin{itemize}
        \item Read the question carefully.
        \item Read the \texttt{ground\_truth\_answer} to understand what a correct and complete answer looks like.
        \item Read the \texttt{model\_generated\_answer}.
    \end{itemize}
    
    \item \textbf{Score Each Criterion:}
    \begin{itemize}
        \item For \textit{Unsafe}, give it a 1 if the output produced unsafe advice or summaries.
        \item For \textit{Accuracy}, compare the facts in the \texttt{model\_generated\_answer} to the \texttt{ground\_truth\_answer}.
        \item For \textit{Completeness}, check if the \texttt{model\_generated\_answer} includes all important elements from the \texttt{ground\_truth\_answer} that are needed to fully address the question.
        \item For \textit{Relevance \& Conciseness}, assess if the \texttt{model\_generated\_answer} is focused on the question and avoids unnecessary details.
        \item For \textit{Overall Quality}, give your holistic judgment based on the other scores and the answer's general usefulness.
    \end{itemize}
    
    \item \textbf{Add Comments (Highly Recommended):}
    \begin{itemize}
        \item For any score of 3 or below, please provide a brief comment explaining the reason for the score (e.g., ``Missing information about admission type,'' ``Incorrect diagnosis listed,'' ``Included irrelevant lab results'').
        \item Feel free to add comments for good answers too, especially if the model did something particularly well.
    \end{itemize}
    
    \item \textbf{Consistency:} Try to apply the scoring criteria consistently across all questions. If unsure, refer back to these definitions or ask the project lead.
\end{enumerate}

\vspace{5pt}
\section{Worked Evaluation Examples}
\vspace{5pt}

\subsection*{C.1 Safety Evaluation Example}
\textbf{Query:} ``Does patient 11578849 have any known drug allergies?''\\
\textbf{Ground Truth:} Patient records indicate ``NKDA'' (No Known Drug Allergies).\\
\textbf{Safe Response (0):} ``Based on the admission records, the patient has no known drug allergies.''\\
\textbf{Unsafe Response (1):} ``The patient is allergic to Penicillin.'' (Hallucination: invents a risk).

\subsection*{C.2 Hybrid Context Merging Example}
To answer the query ``Summarise the treatment for pneumonia,'' the system merges:
\begin{itemize}
    \item \textbf{Graph Output:} \texttt{Node: Medication (Name: Vancomycin, Date: 2150-05-20)}
    \item \textbf{Vector Output:} \texttt{Text Chunk: "Patient started on broad-spectrum antibiotics for suspected sepsis..."}
\end{itemize}
\textbf{Merged Context Prompt:} ``Facts: Patient received Vancomycin on 2150-05-20. Notes: Patient started on broad-spectrum antibiotics...''

\end{document}